\documentclass[conference]{IEEEtran}
\IEEEoverridecommandlockouts
\usepackage{cite}
\usepackage{amsmath,amssymb,amsfonts}
\usepackage{algorithmic}
\usepackage{graphicx}
\usepackage{placeins}
\usepackage{amsmath}
\usepackage{textcomp}
\usepackage{xcolor}
\def\BibTeX{{\rm B\kern-.05em{\sc i\kern-.025em b}\kern-.08em
    T\kern-.1667em\lower.7ex\hbox{E}\kern-.125emX}}
\begin{document}

\title{Edge-Cloud Collaborative Satellite Image Analysis for Efficient Man-Made Structure Recognition  \\
%{\footnotesize \textsuperscript{*}Note: Sub-titles are not captured in Xplore and
%should not be used}
%\thanks{Identify applicable funding agency here. If none, delete this.}
}

\author{
    Kaicheng Sheng\textsuperscript{1,2}, Junxiao Xue\textsuperscript{1,*}, Hui Zhang\textsuperscript{1}\\
    \textsuperscript{1}\textit{Research Center for Space Computing System, Zhejiang Lab, Hangzhou, China}\\
    \textsuperscript{2}\textit{College of Civil and Environmental Engineering, Carnegie Mellon University, Pittsburgh, USA}\\
    kaichens@andrew.cmu.edu, \{xuejx, zhanghui\}@zhejianglab.cn
}
\maketitle

\begin{abstract}
The increasing availability of high-resolution satellite imagery has created immense opportunities for various applications. However, processing and analyzing such vast amounts of data in a timely and accurate manner poses significant challenges. The paper presents a new satellite image processing architecture combining edge and cloud computing to better identify man-made structures against natural landscapes. By employing lightweight models at the edge, the system initially identifies potential man-made structures from satellite imagery. These identified images are then transmitted to the cloud, where a more complex model refines the classification, determining specific types of structures. The primary focus is on the trade-off between latency and accuracy, as efficient models often sacrifice accuracy. We compare this hybrid edge-cloud approach against traditional "bent-pipe" method in virtual environment experiments as well as introduce a practical model and compare its performance with existing lightweight models for edge deployment, focusing on accuracy and latency. The results demonstrate that the edge-cloud collaborative model not only reduces overall latency due to minimized data transmission but also maintains high accuracy, offering substantial improvements over traditional approaches under this scenario.
 
\end{abstract}

\begin{IEEEkeywords}
Satellite image, Edge Computing,  Classification, latency
\end{IEEEkeywords}

\section{Introduction}

The rapid expansion of satellite technologies, particularly those orbiting in Low Earth Orbit (LEO), has greatly transformed our ability to monitor and analyze the Earth's surface. LEO satellites, including constellations operated by entities such as SpaceX, OneWeb, and Telesat, play a pivotal role in a myriad of applications, ranging from global communications to intricate earth observation tasks[1]. These tasks include tracking geomorphic changes, monitoring forest fires, and observing urban expansion dynamics[2], [3]. The data provided by these satellites is invaluable, aiding in everything from climate research to urban planning, and enhancing our response to natural disasters and socio-economical issues. 

Despite their advanced capabilities, most of these satellites still operate under the constraints of the traditional "bent-pipe" architecture, where all captured images are directly transmitted to ground stations for processing and analysis. This traditional method often leads to significant bottlenecks. The limited bandwidth of satellite-ground links, coupled with their unreliability—often intermittent and periodic—can severely disrupt operations. Such disruptions can cause considerable delays in data transmission, critically impairing the timeliness of data analysis and responsiveness[4]. This is particularly detrimental in critical applications such as disaster management and the emerging field of autonomous vehicle navigation, where real-time data is crucial. Take Starlink for example, which is based on bent-pipe, the instability for Starlink across different regions are partly due to the constant network path changes as satellites move closer or further away from the dish, affecting the latency of the single bent-pipe transmission[5].  

Latency, defined as the delay before the transfer of data begins following an instruction, critically undermines the responsiveness of satellite systems[6], particularly in scenarios where time is of the essence. For example, in disaster management, delays in image processing and data transmission can critically hinder the delivery of essential information required to effectively coordinate emergency responses[7]. Such delays can result in slower reaction times during critical events like wildfires or hurricanes, where every second counts towards minimizing damage and organizing effective evacuation plans. latency in self driving is also critical due to the immediate safety implications. For instance, in high-speed driving environments, The accuracy and latency of the systems have great impact on how the vehicle deals with its surroundings[8].

Furthermore, the nature of the data captured by these satellites adds complexity to the task. A significant portion of the Earth's surface imaged by these satellites consists predominantly of natural landscapes. While those images are equally crucial for environmental and scientific studies, the focus of many remote sensing applications is on the detection of human-made structures. This mismatch results in the transmission of vast amounts of data that are not directly relevant to the immediate needs of the tasks at hand, thereby squandering valuable bandwidth and processing resources. To address these inefficiencies, this paper proposes a groundbreaking edge-cloud collaborative architecture that integrates edge computing directly into satellite systems. By incorporating lightweight machine learning models for initial data processing on the satellite, our system can effectively pre-filter and prioritize the transmission of images that contain potential human-made structures. This approach significantly reduces the volume of data requiring transmission and subsequent processing in the cloud, thereby optimizing both bandwidth usage and processing time. This method not only enhances the efficiency of data handling but also ensures that resources are allocated to processing data that offers the most value for real-time decision-making and analysis. Furthermore, The methodology described in this paper is applicable to scenarios where not all satellite images are equally relevant to the task at hand. This approach provides a strategy for enhancing processing efficiency by prioritizing images based on their importance to specific operational goals. 
\section{Background}
\subsection{Edge Computing and Lightweight Model }
Edge computing has emerged in recent years and is being applied to various fields, representing a significant shift in data processing pattern, positioning itself at the frontier of network architecture[9]. This innovative approach involves analyzing and processing data at the periphery of the network, or "the edge," where data is captured and lightweight models are deployed. Unlike traditional centralized computing pattern, which relies on data being sent to remote servers for processing, edge computing brings computational resources closer to the data sources—be it IoT devices, industrial machinery, or mobile devices. This proximity reduces the latency and bandwidth issues often associated with data transfer to central servers, allowing for near-instantaneous data processing and bringing many benefits. 

In the domain of autonomous vehicles, a study by Lv et al.[10] examines the integration of edge computing with 6G networks to enhance environmental perception capabilities of autonomous vehicles. In industrial automation, a compelling study by Schmidt et al.[11] describes the development of a low-cost, Raspberry Pi-based automation system for hydraulic processes, embodying principles of Industry 4.0. In healthcare, an innovative study by Rajavel et al.[12] presents the IoT-based smart healthcare video surveillance system, enhanced through edge computing. This system, named the Cloud-based Object Tracking and Behavior Identification System (COTBIS), dramatically improves fall detection accuracy and reduces response times by implementing advanced video processing algorithms at the network edge.

\subsection{Related Work}

When applied to satellite image, edge computing differentiate itself from traditional computational methods by addressing unique challenges posed by the satellite environment. In satellite systems, particularly those in Low Earth Orbit (LEO), data transmission to ground stations is often hampered by latency, intermittent connections, and limited bandwidth, Moreover, satellites inherently possess constrained computing power and energy resources, necessitating highly efficient data processing solutions that minimize power consumption and computational load. 

One innovative application of edge computing in satellites is the work by Trong-An Bui et al.[13], which leverages an edge-computing-enabled deep learning model for enhancing low-light image processing on satellite platforms. This model employs a lightweight encoder-decoder architecture designed to perform under the constrained computational resources available on satellites. Yijie Chen[14]investigates how to optimize the energy and computational demands that burden traditional DNN applications in satellite operations. Chen proposes a method where DNN tasks are strategically segmented and distributed between satellite platforms and ground stations. This approach not only minimizes latency and power consumption but also enhances the overall efficiency of data processing, making real-time analysis and decision-making feasible even in complex satellite network scenarios. Ming Zhao[15] highlights the integration of collaborative machine learning with satellite operations to enhance the intelligence and responsiveness of satellite systems. Zhao's framework uses asynchronous learning algorithms to accommodate the dynamic and delay-sensitive nature of satellite communication. Bradley Denby[16] explores the transformative potential of Orbital Edge Computing (OEC) within nanosatellite constellations for remote sensing applications. Employing advanced computational modules like the Jetson TX2, Denby's work showcases how edge computing can be implemented directly on satellites to perform sophisticated machine inference tasks onboard. This capability is particularly beneficial for applications requiring immediate data processing, such as environmental monitoring and crisis response, where the speed of data acquisition and analysis is critical. The OEC framework significantly reduces the dependency on ground-based processing infrastructure, thereby decreasing the latency and increasing the timeliness of satellite data utilization. 
\section{\vspace{-0.2cm}System Model}
\subsection{Dataset}\label{AA}
The UCMerced Land Use Dataset was used for this paper. The dataset consists of 2100 aerial scene images, which are categorized into 21 land use classes. The 21 classes include a variety of natural or human-made scenes such as agricultural lands, forest, golf courses, overpasses, parking lots, and residential areas. Each class contains 100 images, offering a balanced dataset that facilitates comprehensive training and testing of machine learning models. The images are uniformly sized at 256x256 pixels, captured at a resolution of 30 centimeters, providing high-detail visual information that is suitable for fine-grained classification tasks. 

\subsection{Methodology}

In this study, we propose a hybrid edge-cloud collaborative architecture for efficient recognition and classification of human-made structures in satellite imagery, as shown in Fig 1. Our methodology consists of initially training a sophisticated model on a local machine using high-performance computing resources. This model focuses on the preliminary identification of images containing man-made structures amidst natural landscapes. A lightweight model is deployed at the edge to handle incoming satellite images, facilitating real-time data processing and immediate decision-making. 

\begin{figure}[h]
\centerline{\includegraphics[width=1\linewidth]{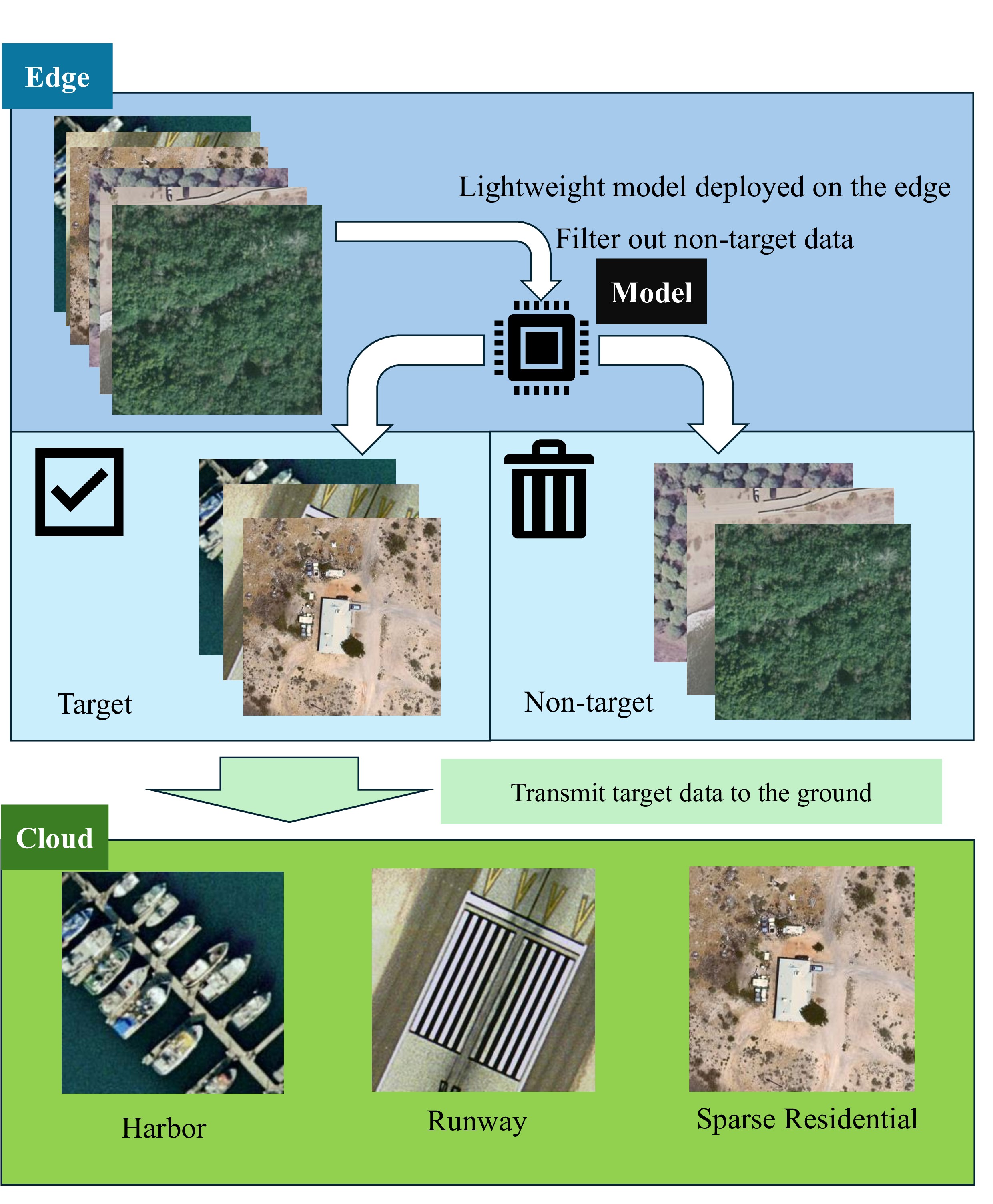}}
\caption{Pipeline of proposed method}
\label{fig}
\end{figure}

At the edge, a lightweight model processes this dataset to swiftly segregate natural landscapes from potential man-made structures, thereby streamlining the dataset to focus solely on images relevant to our task. This filtration significantly alleviates the computational burden and reduces the bandwidth required for transmitting data to the cloud. Once the edge system identifies images containing man-made structures, these are forwarded to the cloud for further analysis. On the cloud, a more sophisticated and computationally intensive model undertakes the task of detailed classification, discerning specific types of man-made structures such as harbor, runway, or sparse residential area. This stage leverages the cloud's enhanced processing power to conduct a thorough and precise analysis, although it incurs a higher time cost due to the depth of processing involved. This hybrid processing framework optimizes the use of distributed computing resources, effectively balancing speed and accuracy in the classification of complex images. The classification of specific types of human-made structures is an important component of the pipeline; however, it falls outside the scope of this paper and will not be addressed in the subsequent sections. 

Various combinations of models were deployed on edge to explore the optimal performance. Neumann, Maxim et al[17] achieved an accuracy of 0.9961 on UCMerced Dataset using ResNet50. MobileNet[18] and ShuffleNet[19] provide an efficient and effective framework for running deep learning models on mobile and edge devices. For every model, we conducted 10 experiments and take the average to avoid the interruption of random netspeed.

Here's a brief introduction about the models we tried in the experiment. SimpleCNN is a self-built convolutional neural network featuring three convolutional layers. Each layer is followed by batch normalization and max pooling to enhance model stability and reduce overfitting.  MobileNetV2 is designed with specific adjustments to its layer configurations and structures, optimizing performance while reducing computational demands and improving accuracy for targeted tasks. We tried ShuffleNet, an efficient convolutional neural network known for its use of pointwise group convolutions and channel shuffle operations. These features significantly reduce the model's complexity and computational cost, making it ideal for applications where resources are limited, without compromising on accuracy. 
\begin{figure*}[ht]
    \centering
    \includegraphics[width=.9\textwidth]{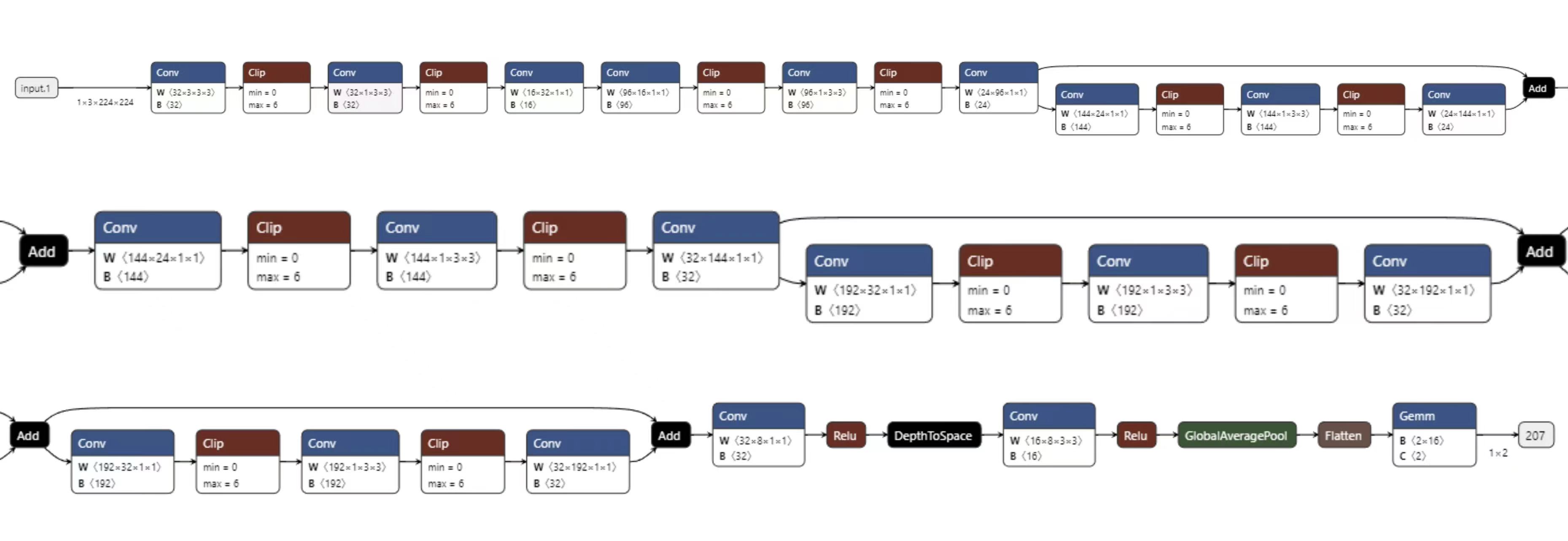}
    \caption{Architecture of MSNet}
    \label{fig:msnet-architecture}
\end{figure*}

The proposed model of this paper, MobileShuffleNet  (MSNet, Fig 2), is an advanced, compact deep learning architecture, particularly suitable for scenarios with limited processing capabilities like handheld devices or on-the-edge computing platforms. In designing the hybrid network architecture, our primary objective was to harness the strengths of both MobileNetV2 and ShuffleNet to create a network that is both lightweight and maintains high performance. MobileNetV2 is renowned for its efficient architecture that employs depthwise separable convolutions to significantly reduce the model's parameter count and computational complexity. On the other hand, ShuffleNet optimizes performance through the use of grouped convolutions and channel shuffling, enhancing both efficiency and information flow.

Initially, the model utilizes the initial layers of MobileNetV2 for foundational feature extraction, predominantly leveraging depthwise separable convolutions for efficient feature processing. Depthwise separable convolutions reduce computational cost by separating the convolution into a depthwise spatial convolution and a pointwise convolution, which can be formally expressed as:
\begin{align}
\text{Depthwise convolution:} \quad Y_k &= X \ast k_k \label{eq:depthwise} \\
\text{Pointwise convolution:} \quad Z &= Y \ast K \label{eq:pointwise}
\end{align}

where \(X\) is the input, \(k_k\) represents the channel-wise convolution kernels, \(K\) is the \(1*1\) convolution kernel, \(Y_k\) denotes the output of the depthwise convolution, and \(Z\) is the final output. This approach drastically cuts down the number of parameters and the computational burden. 

Subsequently, the model incorporates key technologies from ShuffleNet, namely grouped convolutions and channel shuffling. Grouped convolutions divide the input channels into multiple groups, each performing convolutions independently, which not only reduces the parameters further but also lessens the computational load. Channel shuffling ensures effective inter-group information exchange to prevent the isolation of information within groups. Specifically, after each group convolution, a permutation operation rearranges the output feature maps to maintain the network's expressive capability and generalization.The mathematical representation of grouped convolutions can be described as follows: 
\begin{align}
\quad G_i &= \text{GroupedConv}(X, W_i) \quad \text{for } i = 1, \ldots, G \label{eq:grouped_conv}
\end{align}
where \(X\) is the input feature map, \(W_i\) represents the convolutional filters for the i-th group, and \(G\) is the number of groups. 

Channel shuffling can be mathematically expressed as: 
\begin{align}
\quad Y &= \text{Shuffle}(G_1, \ldots, G_G) \label{eq:channel_shuffle}
\end{align}
where \(G_i\),…,\(G_G\) are the grouped convolution outputs before shuffling, and \(Y\) is the shuffled output which combines features from each group. 

To further refine the feature recombination process, a permutation operation rearranges the output feature maps between grouped convolutions:
\begin{align}
\quad R &= \sum_{i=1}^{G} \sigma(W_i \cdot Y_i) \label{eq:feature_combination}
\end{align}
where \(R\) denotes the recombined features, \(W_i\) represents the weights applied to the features of each group \(i\), \(Y_i\) is the output from the channel shuffling phase for each group, and \(\sigma\) is an activation function enhancing non-linearity.

Finally, the network concludes with a simplified classifier comprising global average pooling followed by a fully connected layer. Global average pooling serves as a feature compressor, helping to mitigate overfitting and reduce computational demands, while the fully connected layer caters to task-specific classification. The entire model is designed to strike a balance between computational efficiency and performance, making it suitable for applications that require rapid processing under constrained resources.

\subsection{Implementation details}
 
The experiment is simulated using Ubuntu22.04 virtual environment. Images are resized to a uniform dimension of 64x64 pixels and normalized. The UCMerced dataset is categorized into "Artificial" and "Natural" groups. Artificial classes include structures such as dense residential areas, storage tanks and more. In contrast, natural classes include samples labeled as forests, rivers and others. In this experiment, cross-entropy loss and the Adam optimizer were selected to adjust the model's weights. The training consisted of 25 epochs, with an initial learning rate set at 0.001.

\subsection{Evaluation}
 We meticulously assess both the accuracy and latency of our edge-cloud collaborative model to discern its effectiveness in processing satellite imagery for the detection of man-made structures. Our evaluation is structured to provide a holistic view of the model's performance under operational conditions, focusing on two primary metrics: accuracy and latency. 

In our evaluation of latency, we dissect the time taken at two critical stages to ensure a comprehensive understanding of the system's performance. The first stage, edge processing time, refers to the time required by the satellite's onboard processing system to conduct initial image analysis and classification. This includes any preliminary computations such as feature extraction and basic classification that occur before the data is sent to Earth. The next phase, Transmission Time, is the period needed to transmit the processed data or specific segments of interest from the satellite back to ground stations. This stage is vital as it directly affects how quickly the data becomes available for subsequent analysis, impacting the overall responsiveness of the system. We simulated the weak internet environment typically encountered in space by introducing obstacles to data transmission within an Ubuntu 22.04 virtual machine. Each of these procedures collectively contributes to the total latency of the system, and understanding these components helps in optimizing the pipeline for quicker and more efficient satellite data processing.

\subsection{Results}
\begin{table*}[ht]
    \centering
    \normalsize % 设置大号字体
    \caption{Experiment Results}
    \label{tab:result}
    \begin{tabular}{|c|c|c|c|c|c|} \hline 
         & Bent Pipe & SimpleCNN & MobileNetV2 & ShuffleNet & \textbf{MobileShuffleNet} \\ \hline 
         Edge Processing Time (s) & 0 & 0.81 & 1.18 & 1.02 & \textbf{0.64} \\ \hline 
         Transmission Time (s) & 3.96 & 2.66 & 2.65 & 2.65 & \textbf{2.61} \\ \hline
         Total Time(s)& 3.96 & 3.47 & 3.83 & 3.67 & \textbf{3.25} \\ \hline
         Recall & / & 91.10\% & 97.81\% & 94.41\% & \textbf{97.24\%} \\ \hline 
         Precision & / & 95.48\% & 98.33\% & 97.62\% & \textbf{97.62\%} \\ \hline 
         F1-Score& / & 0.93 & 0.97 & 0.96 & \textbf{0.97} \\ \hline
         Images Transmitted & 420 & 276 & 282 & 279 & \textbf{272} \\ \hline
    \end{tabular}
\end{table*}

%\FloatBarrier  % 保证所有浮动体在此之前被处理

\begin{figure}[ht]
    \centering
    \includegraphics[width=.9\linewidth]{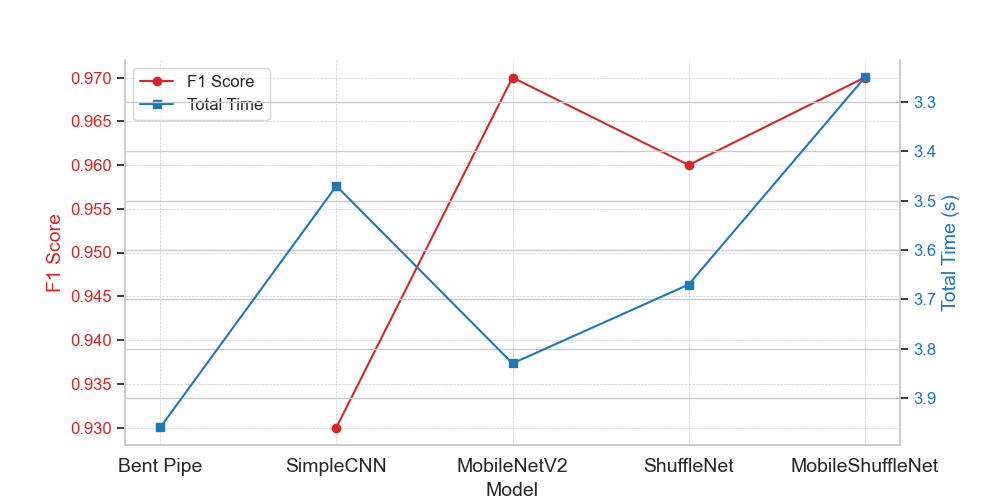}
    \caption{Comparison of F1 Score and Total Processing Time Across Models}
    \label{fig:3}
\end{figure}
\begin{figure}[ht]
    \centering
    \includegraphics[width=.9\linewidth]{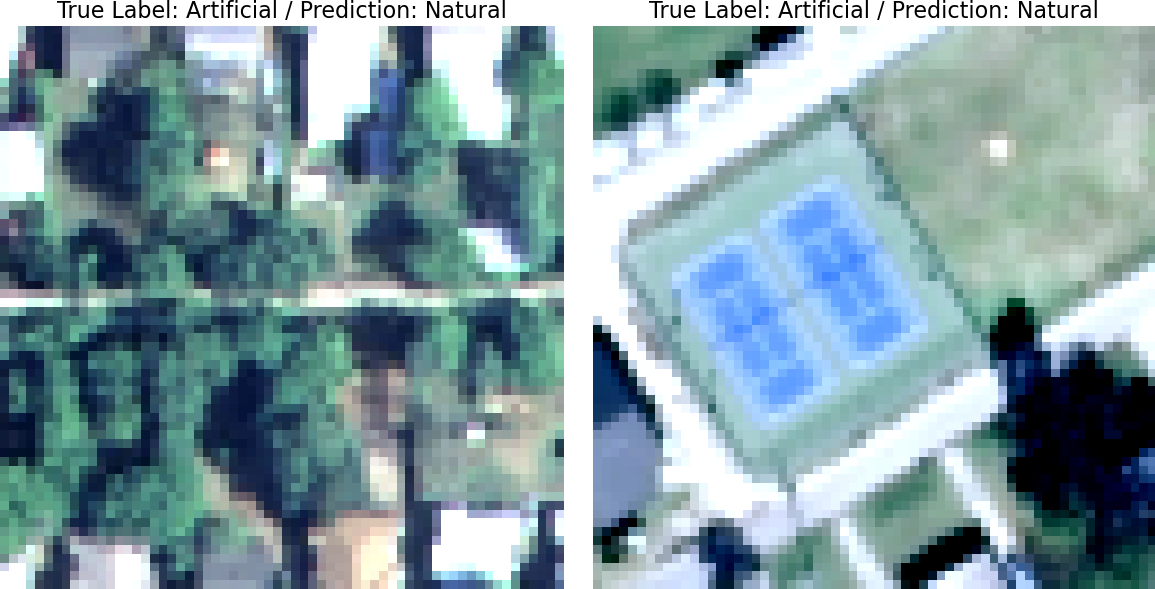}
    \caption{Incorrect predictions}
    \label{fig:4}
\end{figure}
\begin{figure}[ht]
    \centering
    \includegraphics[width=.9\linewidth]{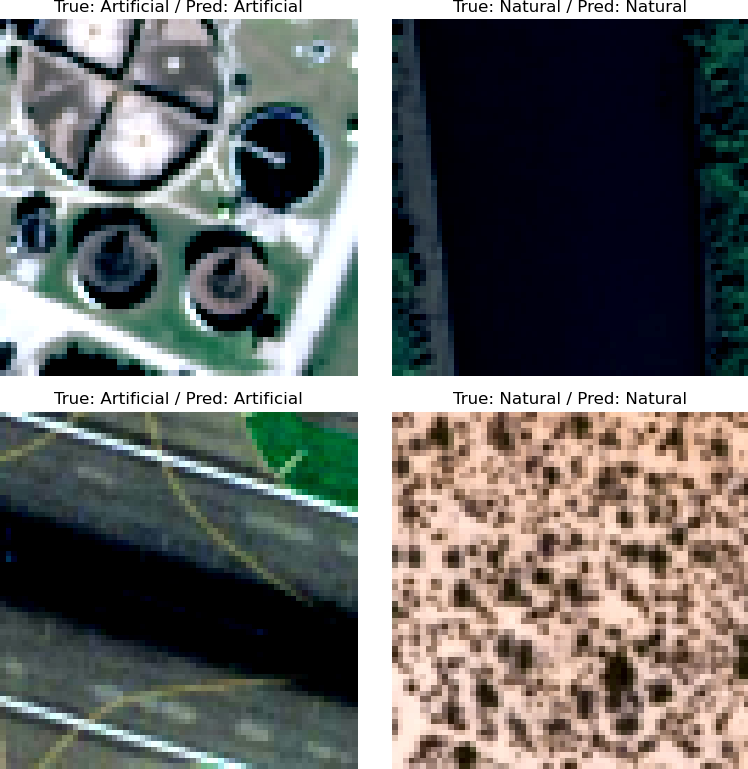}
    \caption{Correct predictions}
    \label{fig:5}
\end{figure}

%\FloatBarrier % 确保所有图形都在这之前显示
The results of the experiment are detailed in Fig 3 and Table 1. Fig 4 and Fig 5 illustrate examples of the prediction alongside the true label. In this study, the edge processing time is influenced by the model's characteristics, while the transmission time depends on the number of images transmitted, a factor also determined by the model. Two key evaluation metrics of interest are F1-score and total time. We found that the proposed method significantly outperforms others. MSNet reduces edge processing time by approximately 37\% compared to ShuffleNet (from 1.02 seconds to 0.64 seconds). The total time required by MSNet is reduced by 11\% compared to ShuffleNet (from 3.67 seconds to 3.25 seconds). When it comes to accuracy, MSNet remains a high F1-score of 0.97, basically same as MobileNetV2 and a bit higher than ShuffleNet.

However, as shown in Fig 4, although MSNet achieves a high F1-score, there are still some incorrect predictions. These errors could lead to severe consequences in practical applications due to their potential for critical oversights. Further studies could be conducted to mitigate these issues.

\section{Conclusion}

This paper presents a novel edge-cloud collaborative architecture as well as a creative model(MSNet) for processing satellite imagery, focusing on the efficient detection and classification of man-made structures against natural landscapes. By employing lightweight models at the edge, the system significantly reduces latency and bandwidth usage by pre-filtering images before transmission to the cloud.  Our evaluation reveals that this hybrid approach not only minimizes overall latency compared to traditional "bent-pipe" methods and some novel lightweight models, but also maintains high accuracy. The results underscore the potential of integrating edge computing with cloud-based analytics to enhance real-time satellite image processing, especially in 6G-integrated satellite networks. This methodology offers a substantial improvement in operational efficiency and responsiveness, making it a valuable contribution to the field of satellite image analysis and real-time decision-making.

\vspace{12pt}

\end{document}